\documentclass[letterpaper, 10 pt, conference]{ieeeconf}  

\IEEEoverridecommandlockouts                              
\usepackage[bookmarks=true]{hyperref}
\usepackage{multicol}
\usepackage[bookmarks=true]{hyperref}

\usepackage{amsmath}
\usepackage{amssymb}
\usepackage{subcaption}
\usepackage{graphicx}
\usepackage{xcolor}
\usepackage{makecell}
\usepackage{hhline}
\usepackage[]{algorithm2e}
\usepackage[super]{nth}
\usepackage{siunitx}  

\newcommand\norm[1]{\left\lVert#1\right\rVert}

\DeclareMathOperator*{\argmin}{argmin}   

\title{\LARGE \bf
Control-Tree Optimization: an approach to MPC under discrete Partial Observability
}

\author{Camille Phiquepal$^{1}$ and Marc Toussaint$^{2}$
\thanks{$^{1}$Machine Learning \& Robotic Lab, University of Stuttgart, Germany
        {\tt\small camille.phiquepal@ipvs.uni-stuttgart.de}}%
\thanks{$^{2}$Learning and Intelligent Systems Lab, TU Berlin, Germany
        {\tt\small toussaint@tu-berlin.de}}%
}

\begin{document}


\maketitle
\thispagestyle{empty}
\pagestyle{empty}

\begin{abstract}

This paper presents a new approach to Model Predictive Control for environments where essential, discrete variables are partially observed. Under this assumption, the belief state is a probability distribution over a finite number of states.
We optimize a \textit{control-tree} where each branch assumes a given state-hypothesis. The control-tree optimization uses the probabilistic belief state information. This leads to policies more optimized with respect to likely states than unlikely ones, while still guaranteeing robust constraint satisfaction at all times. We apply the method to both linear and non-linear MPC with constraints. The optimization of the \textit{control-tree} is decomposed into optimization subproblems that are solved in parallel leading to good scalability for high number of state-hypotheses. We demonstrate the real-time feasibility of the algorithm on two examples and show the benefits compared to a classical MPC scheme optimizing w.r.t. one single hypothesis.

\end{abstract}

\section{INTRODUCTION}


In the field of receding horizon motion planning, uncertainty about the robot environment is often neglected or integrated in the environment representation through heuristics (e.g. by adding safety distances, or potential fields around perceived objects). At each planning cycle one single environment representation is used for planning the next motion. Reactivity and robustness are achieved by re-planning. 

In contrast, we aim here to tackle problems where the uncertainty is about discrete and critical aspects of the environment, such that the agent has to anticipate for multiple hypothetical scenarios within the planning horizon.

The proposed approach is inspired from \cite{tamp-1} where an integrated Task and Motion planner is introduced for partial observable environments. \cite{tamp-1} focuses mainly on off-line planning for object manipulation and optimizes trajectory-trees that react to observation. This paper transfers this concept to Model Predictive Control and real-time trajectory optimization.
Accordingly, the main contributions of this paper are:
\begin{itemize}
\item An MPC formulation for optimizing control policies (\textit{control-trees}) under discrete partially observability. It leverages the probabilistic information of the belief state to reduce conservativeness.
\item An algorithm \textit{Distributed Augmented Lagragian} combining the Augmented Lagrangian and ADMM method. It allows fast and distributed optimization of the control-tree, is applicable to both linear and non-linear MPC, and scales linearly w.r.t. the number of state hypotheses.
\item The demonstration of the benefits and real-time feasibility of the method on robotics problems.
\end{itemize} 

%
%

\section{RELATED WORK}



Stochastic and Robust variants of Model Predictive Control aim at optimizing performance and constraint satisfaction under disturbances (e.g. un-modeled system dynamics, external disturbances, noise). Good surveys on Stochastic and Robust Optimal can be found in \cite{soc-1} and \cite{soc-2}. Robust MPC seeks constraints satisfaction at all times by tackling the worst cases of the uncertainty. In contrast, Stochastic MPC uses the information how the uncertainty is distributed and interprets constraints probabilistically: a constraint is satisfied if the probability of a violation is kept below some threshold. Like Robust MPC, the presented approach seeks constraint satisfaction even in the worst case. As per Stochastic MPC, it leverages the stochastic information to reduce the conservativeness.



Several approaches have been developed that consider multiple scenarios within the planning horizon.

Min-Max MPC \cite{robust-1}\cite{min-max-1} can be considered as one of these, which optimizes a control sequence w.r.t. the worst case. Although this guarantees that constraints are met at all times, it can be over-conservative.

 


Approaches optimizing control-trees have already been used in MPC, mainly for optimal control of chemical processes. In \cite{multi-stage-luca-1-LUCIA201269}\cite{multi-stage-luca-1-LUCIA20131306}\cite{multi-stage-luca-1} Lucia and al. introduced a tree-based approach called \textit{multi-stage} MPC. This has been developed further, e.g. Kouzoupis and al. in \cite{multi-stage-klintberg-1}, Leidereiter and al. \cite{multi-stage-leidereiter-1}.
 
To our knowledge \cite{multi-stage-robotics-1} and \cite{multi-stage-klintberg-2} are the only published experiments applying \textit{multi-stage} MPC to mobile robotic use-cases.
In all those previous applications of multi-stage MPC, scenarios are used to model uncertainty over continuous variables (by enumerating possible realizations). Its application as a treatment of partial observability, with intergation of the belief state is novel. Moreover, unlike \cite{multi-stage-klintberg-2}, the approach is not restricted to linear MPC, and compared to \cite{multi-stage-robotics-1}, we add an optimization algorithm allowing distributed computation to scale to high number of hypotheses.

\section{Problem formulation}
\subsection{Overview}
\begin{figure}[ht]
\begin{subfigure}{.15\textwidth}
  \centering
  \includegraphics[width=0.95\linewidth]{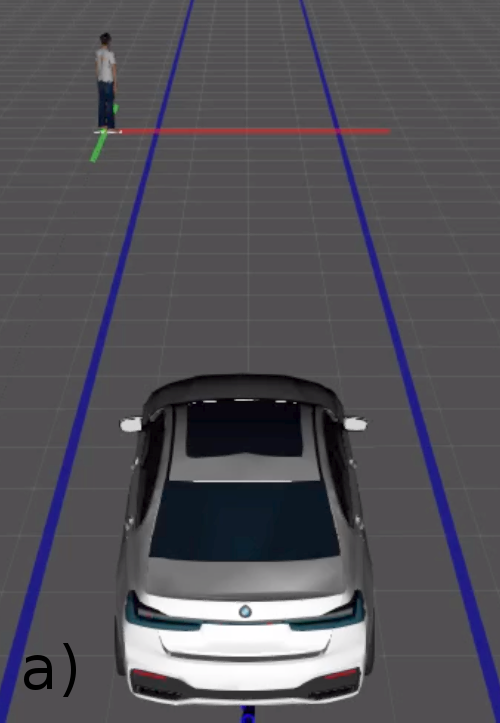}  
\end{subfigure}
\begin{subfigure}{.15\textwidth}
  \centering
  \includegraphics[width=.95\linewidth]{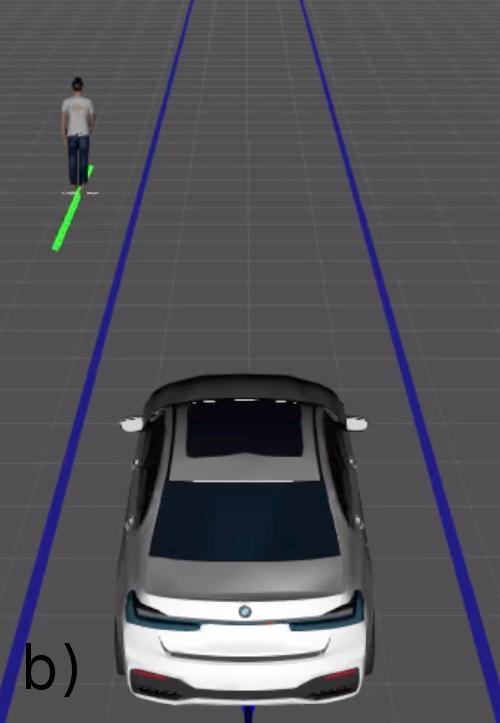}  
\end{subfigure}
\begin{subfigure}{.15\textwidth}
  \centering
  \includegraphics[width=.95\linewidth]{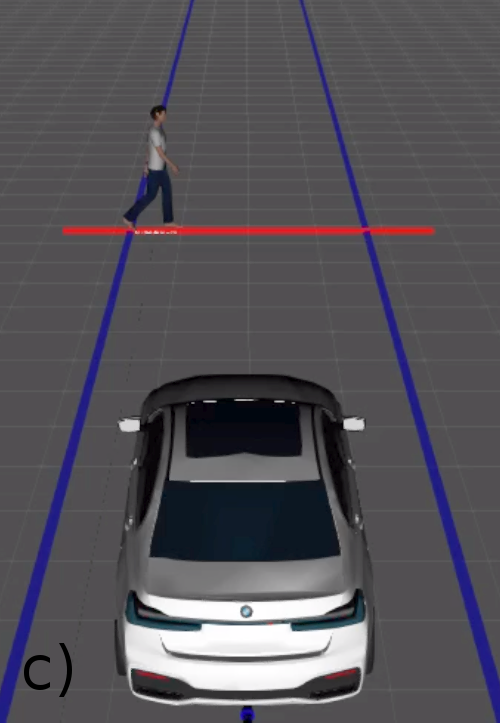}  
\end{subfigure}
\caption{In a) The pedestrian is standing, his intention is uncertain. He may walk along the street b) or cross c). It is modeled as a partially observable discrete state.}
\label{fig:example}
\end{figure}

The presented approach optimizes control policies in a context of mixed-observability. The usual continuous state variables (e.g. position, velocity) are assumed fully observed. On the other hand, discrete, essential parts of the environment are partially observed, this is modelled by a discrete state. In the example of Figure \ref{fig:example}), this models the pedestrian intention. 

We assume that the agent is not oblivious to the likelihood of each hypothesis: it maintains a belief state, i.e. a probability distribution over the multiple hypotheses. The belief state is used to optimize the policy more w.r.t. likely states.

\begin{figure}[!htb]
 \center{\includegraphics[width=0.47\textwidth]{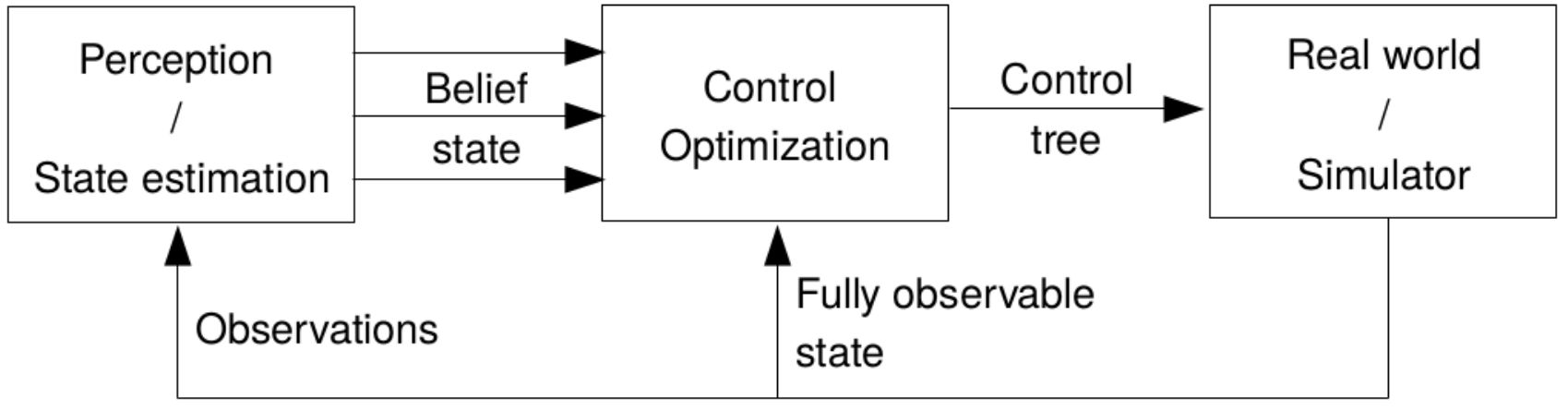}}
 \caption{Control loop: control-trees are optimized w.r.t. a belief state (probability distribution over multiple hypotheses).}
 \label{fig:control-loop}
\end{figure}

\subsection{State representation}
We consider a compound state representation where a state is composed of 2 parts:
\begin{itemize}
\item $x \in \mathbb{R}^n$: system continuous state
\item $s \in S$: discrete state from a finite state space $S$
\end{itemize}

The continuous state $x$ corresponds to the classical notion of state in the MPC literature, and is assumed fully observable. 

The discrete state $s$ represents discrete aspects of environment and is only partially observed.



\subsection{Control-tree}
\begin{figure}[!htb]
 \center{\includegraphics[width=0.4\textwidth]{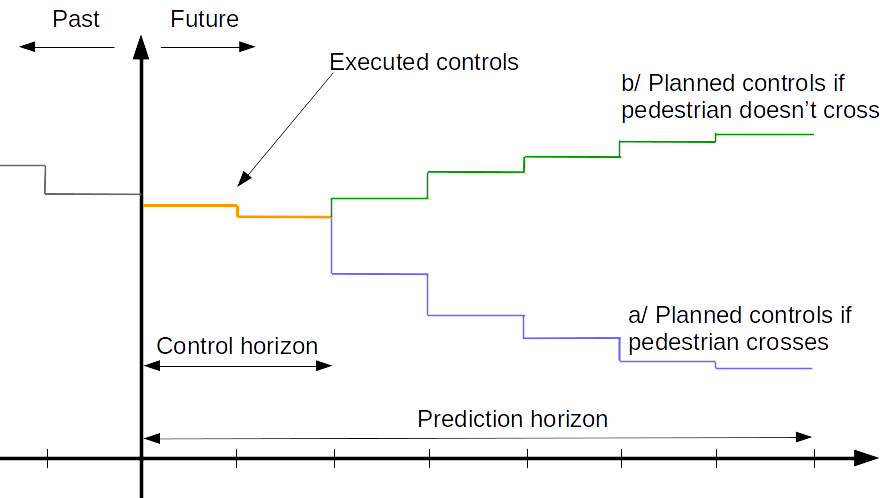}}
 \caption{\label{fig:control-tree} Schematic control-tree applied to the pedestrian example: After the control horizon, the control-tree evolves in two different branches corresponding to the two different hypothetical states (pedestrian crossing or not).}
\end{figure}

The algorithm consists in optimizing a control-tree. The control-tree has a first common trunk (see orange part in Figure \ref{fig:control-tree}). This part is executed by the controller until the next planning cycle happens. We call \textit{control horizon} the time interval corresponding to this common trunk.
 After the control horizon, the control-tree branches, and each branch is optimized assuming a certain state $s$.
 This is akin to the QMDP method (see \cite{qmdp}) for solving POMDPs, but here in continuous domain.

\subsection{Optimization problem formulation}
We note $L$ the number of steps in the control horizon, and $T$ the total number of steps on each branch ($ L \ll T$).

\noindent Let $u_s$ and $x_s$, $s \in S$ be the control and configuration sequences on the branch corresponding to the state $s$. Let $\tilde{u}$ be the controls in the control horizon. Let $p(s)$ be the probability of state $s$.

\noindent We formulate the optimal control problem as follows:
\begin{subequations} \label{eq:optimization-problem}
\begin{align}  
  \min_{\tilde{u}, u_s, x_s}&{
  \sum_{s \in S}
  {p(s)\sum_{t=0}^{T-1}
  {
  c_s(x_s(t), u_s(t)) \label{eq:cost-term}
  }
  }
  },\\
  &\text{s.t.} \notag \\
  & g_s(x_s(t), u_s(t)) \leq 0,&&\forall{s}, \forall{t}\label{eq:constraint-term},\\
  & x_s(t+1) = f(x_s(t), u_s(t)),&&\forall{s}, \forall{t}\label{eq:mpc_model},\\
  & u_s(t) = \tilde{u}(t),&& \forall{s}, \forall{t < L},\label{eq:non-anticipativity}
\end{align}
\end{subequations}
where $c_s$ and $g_s$ are the cost and inequality constraints applying in state $s$ respectively. The function $f$ models the system dynamic. 

The optimization variables $u_s$ and $x_s$, $s \in S$ are not fully independent from each other. Indeed, in the control horizon, they all correspond to the common trunk and must therefore be equal. This is captured by the equation (\ref{eq:non-anticipativity}) and is usually called "non-anticipativity" constraint in multi-stage MPC \cite{multi-stage-luca-1-LUCIA201269}. In the following we call $\tilde{u}$ the consensus variable.

The common trunk of the tree is constrained by the active constraints of all states (see (\ref{eq:constraint-term})), regardless of the state likelihood. This is for guaranteeing the robustness of the constraint satisfaction.

On the other hand, the minimized costs (\ref{eq:cost-term}) are weighted by the belief state for optimizing more w.r.t. likely states than unlikely ones.\\

\subsection{Transcription to generic solver format} \label{section:problem_reduction}
Here we rewrite the optimization problem in a generic format that is the input to our solver. In Equation (\ref{eq:optimization-problem}) both the controls $u_s$ and states $x_s$ are optimization variables.
In many cases it is possible to eliminate either the controls or the configurations and obtain a more compact formulation:
\begin{itemize}
\item
Optimization in control space. This is achieved by eliminating the $x_s$ and is well described in the MPC literature, in particular for linear MPC (e.g. \cite{Diehl2013}). We apply this scheme in the first experiment in section \ref{section:acc}.
\item 
Optimization in configuration space. This typically requires adding additional constraints to ensure the existence of controls implementing the configuration transitions e.g. for non-holonomic robots. The second experiment in section \ref{section:slalom} follows this approach.
\end{itemize}

The problem can be rewritten in a generic format:
\begin{subequations} \label{eq:gen-optimization-problem}
\begin{align}  
  \min_{\tilde{z}, z_s}&{
  \sum_{s \in S}
  {p(s) c_s(z_s) \label{eq:gen-cost-term}
  }
  },\\
  &\text{s.t.} \notag \\
  & g_s(z_s) \leq 0,&&\forall{s}\label{eq:gen-constraint-term},\\
  & h_s(z_s) = 0,&&\forall{s}\label{eq:gen-mpc_model},\\
  & z_s(t) = \tilde{z}(t),&& \forall{s}, \forall{t < L},\label{eq:gen-non-anticipativity}
\end{align}
\end{subequations}
where $\tilde{z}$ and $z_s, s \in S$ are the optimization variables, which can be in control space, configuration space, or a combination of both.
We note $d$ the dimensionality of the optimized parameters at each time step. For each $s \in S$, the function $c_s:\mathbb{R}^{T \times d} \rightarrow \mathbb{R}$ is a scalar objective function, $g_s:\mathbb{R}^{T \times d} \rightarrow \mathbb{R}^{d_{g_s}}$ and $h_s:\mathbb{R}^{T \times d} \rightarrow \mathbb{R}^{d_{h_s}}$ define $d_{g_s}$ inequality and $d_{h_s}$ equality constraints respectively. (\ref{eq:gen-non-anticipativity}) is the non-anticipativity constraint and $\tilde{z}$ is the consensus variable. We generally assume the functions $c_s$, $g_s$, and $h_s$ to be smooth, but not necessarily convex or unimodal.

\section{Solver}

The global optimization problem (\ref{eq:gen-optimization-problem}) can be decomposed into $N$ loosely coupled optimization problems with $N = |S|$. Indeed the $N$ tree branches are nearly-independent from each other (except for the non-anticipativity constraint (\ref{eq:gen-non-anticipativity})). The core idea of the proposed algorithm is to take advantage of this decomposition.
It performs multiple iterations consisting each of 2 phases:
\begin{itemize} 
\item A distributed phase where a relaxed version of each subproblem is optimized.
\item A centralized phase at which the consensus variable $\tilde{z}$ is updated.
\end{itemize}
 
\subsection{Distributed Augmented Lagrangian} 
The optimization problem of each branch is nearly a standard constrained optimization problem. Only the non-anticipativity constraint makes it peculiar, because the consensus variable $\tilde{z}$ varies over time and depends on the results of the other subproblems. 

To solve this, we form an unconstrained objective function which combines the augmentations of both the Augmented Lagragian method (AuLa) and the Alternating Direction Method of Multipliers (ADMM). We call it the \textit{Distributed Augmented Lagrangian}:
\begin{subequations} \label{eq:aula-problem}
\begin{align}  
L_s(z, \tilde{z}, \lambda_s, \kappa_s, \eta_s) &= p(s) c_s(z) \\
&+ \lambda_s^\intercal g_s(z) + \mu \norm{[g_s(z) > 0] \cdot g_s(z)}^2 \label{eq:lagrangian-ineq}\\ 
&+ \kappa_s^\intercal h_s(z) + \nu \norm{h_s(z)}^2 \label{eq:lagrangian-eq}\\ 
&+ \eta_s^\intercal (\Delta z)  + \frac{\rho}{2} \norm{\Delta z)}^2,\label{eq:admm}
\end{align}
\end{subequations}
where $\Delta z \in \mathbb{R}^{L\times d}$, with $\Delta z(t) = z(t) - \tilde{z}(t), \forall t < L$ is the difference between $z$ and the consensus $\tilde{z}$ in the control horizon. 
 
(\ref{eq:lagrangian-ineq}) and (\ref{eq:lagrangian-eq}) are the terms of the Augmented Lagrangian for handling the inequality and equality constraints intrinsic to each subproblem (see \cite{toussaint2014novel}, \cite{conn}).

(\ref{eq:admm}) are the similar ADMM terms for solving the coupling between the subproblems (see \cite{ADMM}). 

 
$\lambda_s \in \mathbb{R}^{d_{g_s}}, \kappa_s \in \mathbb{R}^{d_{h_s}}, \eta_s \in \mathbb{R}^{L \times d}$ are the dual variables, and $\mu, \nu, \rho$ are fixed real constants.


\subsection{Optimization procedure}
The optimization algorithm is the following:
\begin{subequations} \label{eq:aula-problem}
\begin{align}  
{z_s^{k+1}} &:= \argmin_{z_s} L_s(z^k_s, \tilde{z}^k, \lambda_s^k, \kappa_s^k, \eta_s^k) \label{eq:algo-newton}\\
\lambda_s^{k+1} &:= \max(0, \lambda_s^{k} + 2\mu g_s(z_s^{k+1})) \label{eq:algo-dual-ineq} \\
\kappa_s^{k+1} &:= \kappa_s^{k} + 2\nu h_s(z_s^{k+1}) \label{eq:algo-dual-eq} \\
\tilde{z}^{k+1} &:= \frac{1}{N} \sum_{s \in S}{z_s^{k+1}} \label{eq:z_update} \\
\eta_s^{k+1} &:= \eta_s^{k} + \rho \Delta z_s^{k+1} . \label{eq:dual-admm}
\end{align}
\end{subequations}

\subsubsection{Initialization}
All dual variables $\lambda_s, \kappa_s, \eta_s$ are initially set to $0.0$.
The optimization variables $z_s$ as well as the initial reference $\tilde{z}$ can be initialized randomly, or by any better heuristic to speed-up.

\subsubsection{Unconstrained minimization}
\begin{figure}[!htb]
 \center{\includegraphics[width=0.475\textwidth]{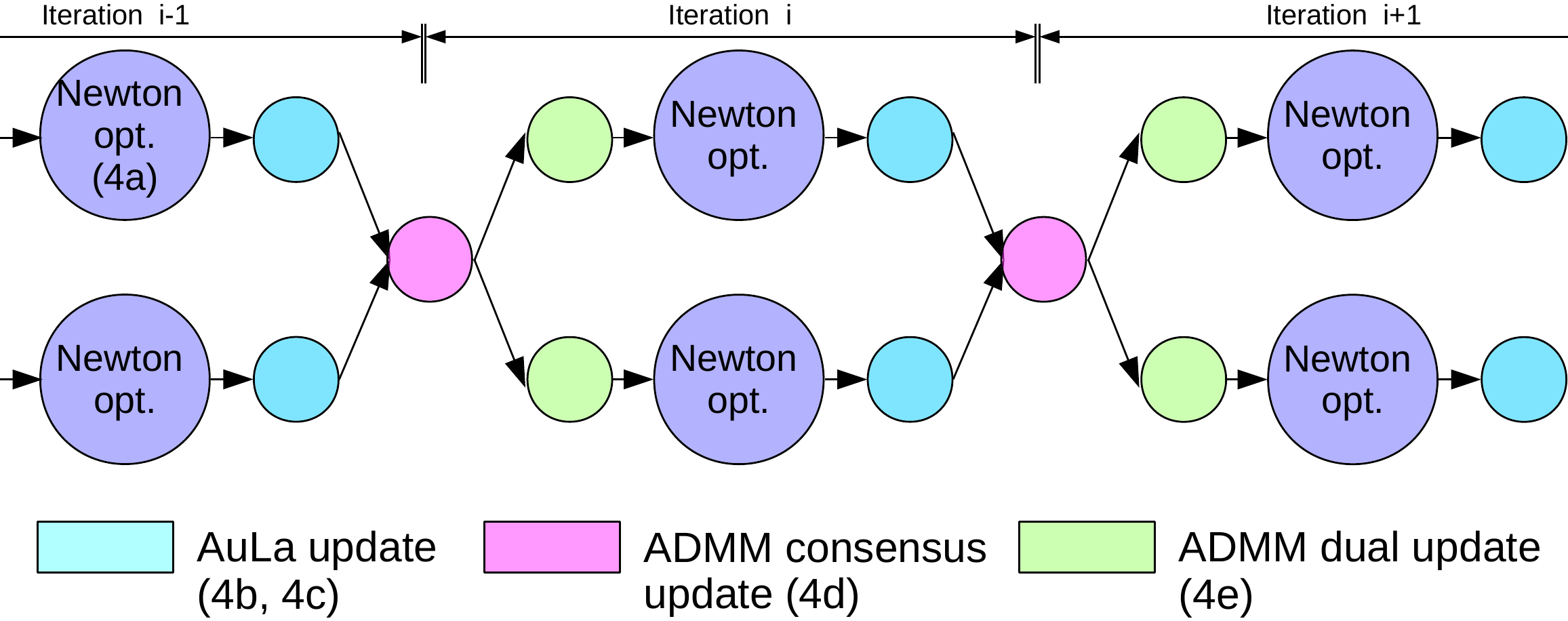}}
 \caption{Execution flow for $N=2$. The costly steps (Newton minimizations) are parallelized.}
 \label{fig:optim-flow}
\end{figure}

The step (\ref{eq:algo-newton}) is the minimization of $N$ unconstrained optimization problems and is the part which is computationally expensive. We optimize them with a newton procedure. As Figure \ref{fig:optim-flow} shows, these optimizations can all be performed in parallel. This is where the algorithm takes full advantage of the decomposition. 


\subsubsection{Augmented Lagrangian dual variables update}
Equations (\ref{eq:algo-dual-ineq}) and (\ref{eq:algo-dual-eq}) update the dual variables corresponding to the inequality and equality constraints respectively.

\subsubsection{ADMM dual and consensus variables update}
Line (\ref{eq:z_update}) updates the consensus variable. It can be performed only when the computations of all the $z_s^{k+1}$ are finished. Its computation is fairly intuitive, it averages the results of each branch over the control horizon (to lighten the notation we omitted the time bound $t<L$ in (\ref{eq:z_update})). The ADMM dual variables are updated in (\ref{eq:admm-dual}). Over the course of the optimization, these dual and consensus  updates ensure that all $z_s$ converge to a same consensus $\tilde{z}$ for $t<L$.

In the literature, the ADMM algorithm usually refers to a decomposition into 2 subproblems that are solved and updated in a sequential fashion. We use here the ADMM variation called \textit{consensus optimization} (see \cite{ADMM}) for a \textit{N}-fold decomposition, and where the optimizations of the subproblems are parallelizable.
\subsubsection{Termination criterion}
The procedure can be stopped once the constraints of each subproblem are satisfied and once a consensus is reached for $t < L$. Formally it means that both the AuLa and ADMM primal and dual residuals are smaller than threshold values ($\epsilon^{pri}, \epsilon^{dual}, \xi^{pri}, \xi^{dual} \in \mathbb{R}$). 


\begin{subequations} \label{eq:aula-problem}
\begin{align}
 &\norm{g_s(z^{k+1})} \leq \epsilon^{pri}, \norm{h_s(z^{k+1})} \leq \epsilon^{pri}, \label{eq:aula-primal} \\ &\norm{z_s^{k+1} - z_s^{k}} \leq \epsilon^{dual}, \label{eq:aula-dual} \\
 &\norm{z_s^{k+1} - \tilde{z}^{k+1}} \leq \xi^{pri}, \label{eq:admm-primal} \\
 &\norm{\tilde{z}^{k+1} - \tilde{z}^{k}} \leq \xi^{dual}. \label{eq:admm-dual}
\end{align}
\end{subequations}

In the examples of the experimental section, the algorithm typically converges after 10 to 30 iterations.

\section{Experiments}
The solver is implemented in C++. The source code and a supplementary video are available for reference\footnotemark \footnotetext{\href{https://github.com/ControlTrees/icra_2021}{https://github.com/ControlTrees/icra2021}}.
\subsection{Adaptative Cruise Control among pedestrians} \label{section:acc}
We consider the problem briefly introduced in Figure \ref{fig:example}. The car drives along a street in presence of pedestrians on the sides who may cross. The pedestrians' intentions are partially observed through a simulated perception module that outputs, for each pedestrian, the probability that it will cross in front of the car. Eventually pedestrians either cross the street, or walk on the walkway making their intention fully observable. 

We optimize the longitudinal acceleration of the car. The car dynamic is modeled as a linear system (\ref{eq:dynamic}) and we consider quadratic cost with linear constraints. The problem is reduced to an optimization in control space only (see section \ref{section:problem_reduction}). (\ref{eq:gen-optimization-problem}) takes the form of $N$ loosely-coupled QPs. 


Performance are evaluated on randomized scenes in simulation. The car dynamic is simulated in Gazebo. Planning occurs at 10Hz with a \SI{5}{\second} horizon and 4 steps per second.

We compare the results obtained with the control-tree optimization vs. a baseline. The baseline is a classical MPC approach considering one single hypothesis. To make sure that no collision happens with pedestrians, the hypothesis used is the worst case: as long as the pedestrian's intention is uncertain, the car plans to stop in front of it. 


\subsubsection{System dynamic}
Let $x$ be the longitudinal coordinate of the ego-vehicle along the road.
The system dynamic is described by the linear system:
\begin{align}
\begin{pmatrix} 
x_{t+1} \\
 v_{t+1}
\end{pmatrix} 
= 
\begin{pmatrix}
 1 & dt \\
 0 & 1
\end{pmatrix}
\begin{pmatrix} 
x_{t} \\
 v_{t}
\end{pmatrix} 
+
\begin{pmatrix} 
0 \\
dt
\end{pmatrix}
\begin{pmatrix} 
u_t
\end{pmatrix}\ , \label{eq:dynamic}
\end{align}
 where $v_t$ and $u_t$ are respectively the vehicle speed and the controlled acceleration. 

\subsubsection{Partially observable discrete state}
With $N_p$ pedestrians located at $x_0< ..<x_{N_p-1}$ ahead of the vehicle, the discrete state can be described by an integer $s \in [0..N_p]$ indicating the closest pedestrian who crosses. $s=N_p$ is the case where no pedestrian crosses.

We note $p_i$ the output of the simulated prediction module giving the crossing probability of the $i^{th}$ pedestrian. The $s^{th}$ pedestrian is the closest crossing if it crosses, and, the pedestrians before him don't cross, such that:
\begin{align}
p(s) = p_s\prod_{i=0}^{s-1}1-p_i \ .
\end{align}

\subsubsection{Costs and constraints}
The control-tree is optimized w.r.t the following trajectory costs:
\begin{itemize}
\item \textbf{Speed}: The matrix 
$\boldsymbol{Q}=\begin{pmatrix} 
0 & 0\\
0 & k_v \footnotemark
\end{pmatrix}$ penalizes the velocity difference between the vehicle speed and a given desired velocity.
\item \textbf{Acceleration}:
The square acceleration is penalized, $\boldsymbol{R}=\begin{pmatrix} k_u \footnotemark[\value{footnote}] \end{pmatrix}$.
\end{itemize}
In addition, the following constraints are applied:
\begin{itemize}
\item \textbf{Stop before the $i^{th}$ pedestrian}: A state inequality constraint applies to stop and keep a safety distance to the pedestrian, $x \leq x_{i} - d_{saftey} \footnotemark[\value{footnote}]$.
\item \textbf{Control bounds}: Longitudinal acceleration is constrained to stay between bounds $[-8.0, 2.0] m/{s^2}$.
\end{itemize}

\footnotetext{Table \ref{tab:acc_costs} is obtained with $k_u = 5.0$, $k_v = 1.0$, $d_{safety} = 2.5$ m}

\subsubsection{Example of control-trees}
\begin{figure}[!htb]
 \center{\includegraphics[width=0.45\textwidth]{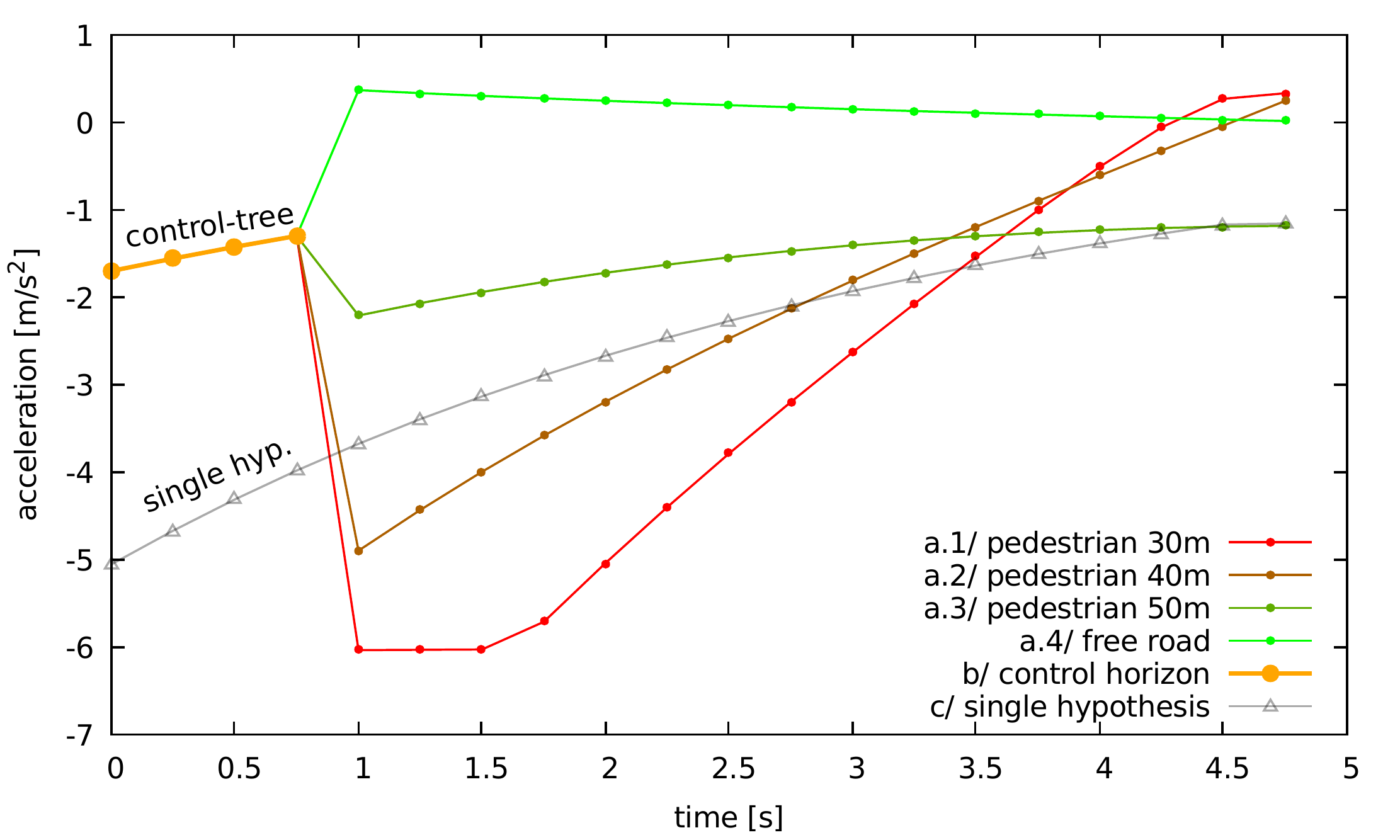}}
  \center{\includegraphics[width=0.45\textwidth]{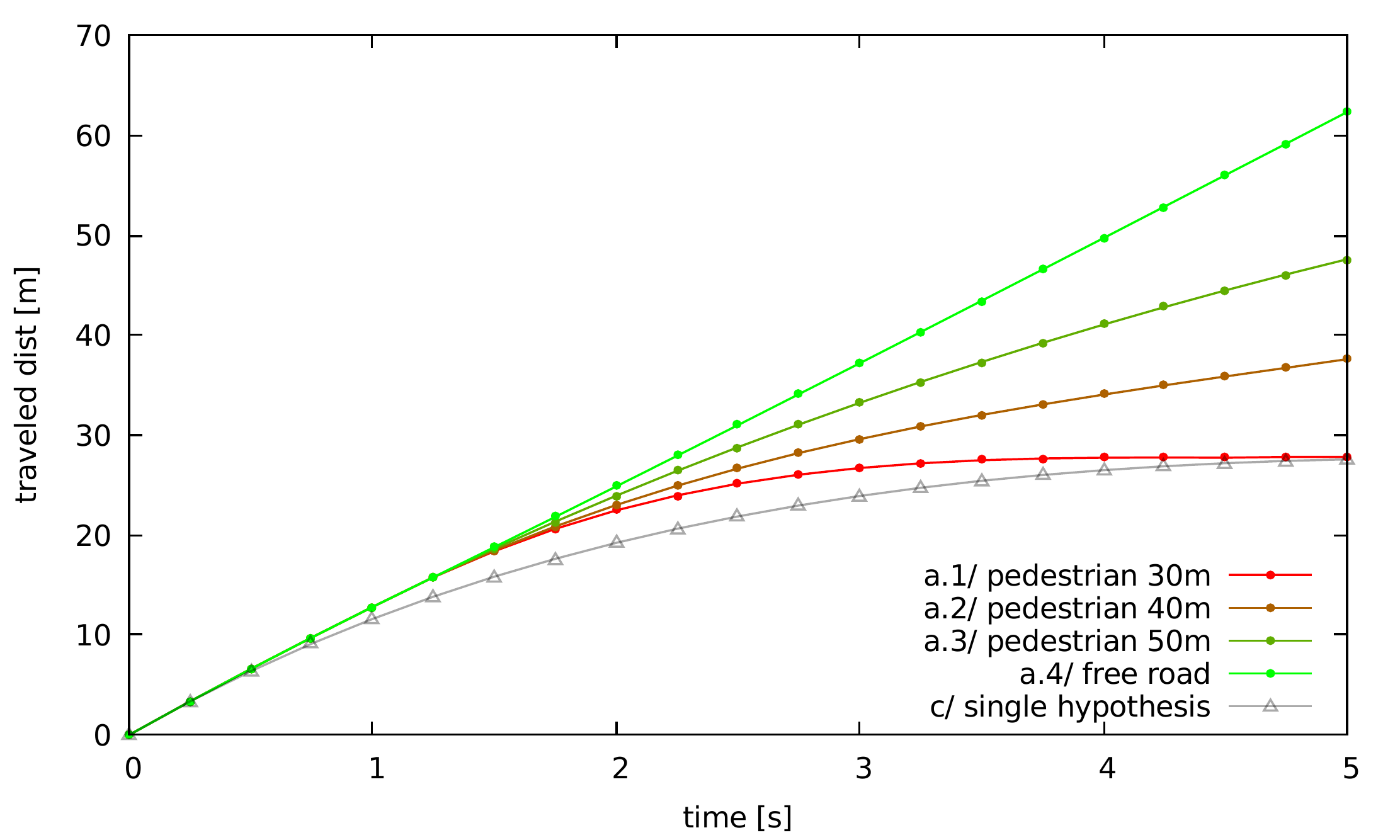}}
  \caption{Example of control-tree when braking: a.1, a.2, and a.3 anticipate that a pedestrian crosses. a.4 corresponds to the free road scenario. c. is obtained with single hypothesis MPC assuming that the closest pedestrian crosses.}
\label{fig:control-tree-vs-sequential}
\end{figure}

Figure \ref{fig:control-tree-vs-sequential} shows a control-tree obtained with a vehicle launched at $48$ km/h ($30$mph) in the presence of 3 pedestrians. Each pedestrian has a probability $0.15$ that it will cross. This implies a probability of $0.61$ that the road is free. The  control-tree doesn't brake too hard, but still guarantees that it is possible to come to a stop in the worst case (see red curve in Figure \ref{fig:control-tree-vs-sequential}). On the other hand, the single hypothesis approach brakes much stronger.

\subsubsection{Influence of the belief state}
\begin{figure}[!htb]
 \center{\includegraphics[width=0.5\textwidth]{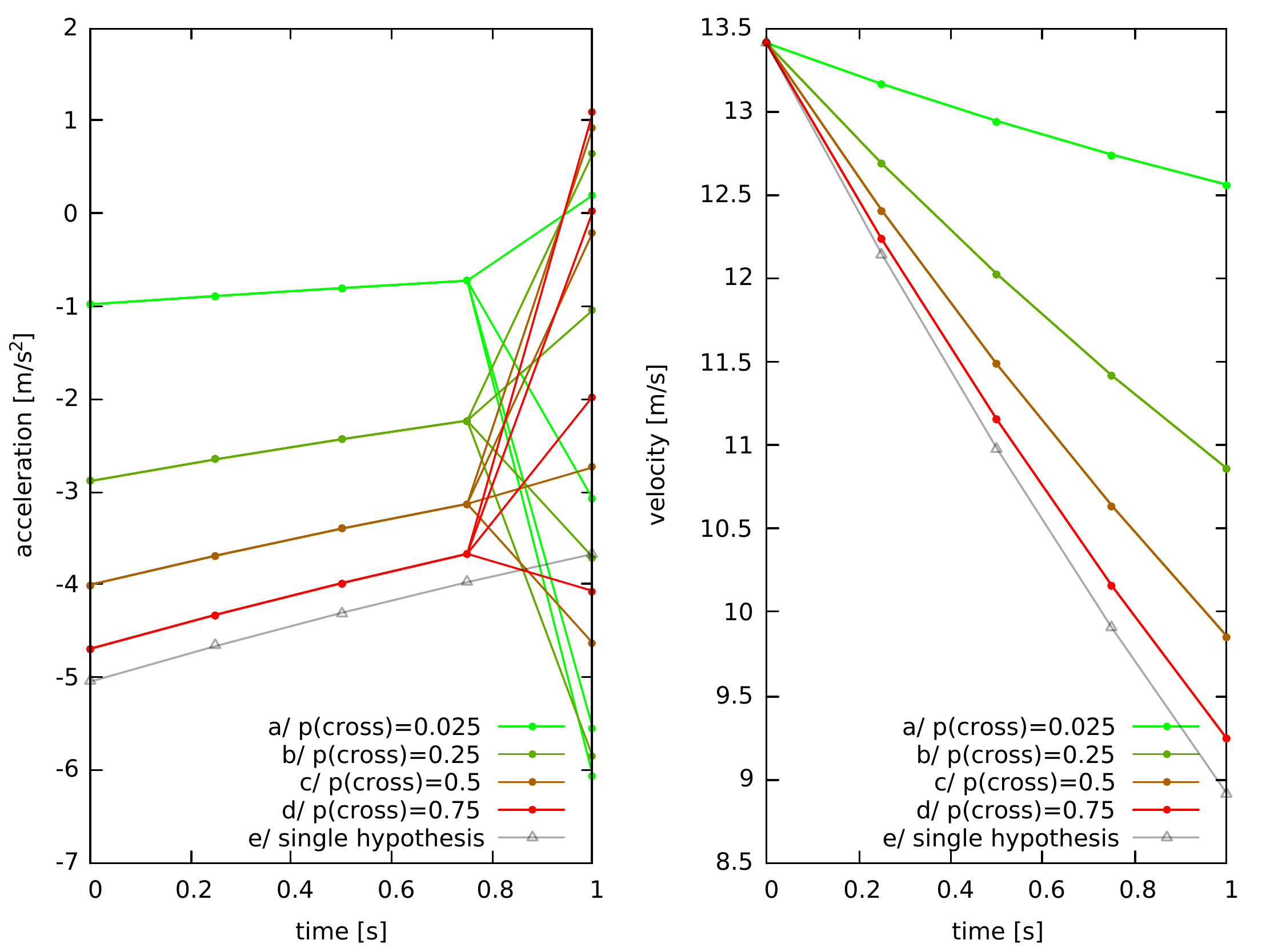}}
  \caption{Influence of the belief state on the braking within the control horizon: Low crossing probabilities (e.g. a, b) lead to a more optimistic control-tree.}
\label{fig:belief-state}
\end{figure}

The Figure \ref{fig:belief-state} focuses on the control horizon ($t\leq 1.0s$) and shows the influence of the crossing probability. When the probability is low, the planned control is more optimistic, whereas when this probability increases, the control policy becomes more conservative and tends to the limit-case (single hypothesis).

\subsubsection{Evaluation on random scenarios}
The algorithm is tested under various combinations of pedestrian density and pedestrian behavior (average crossing probability). Each run is performed over 30 minutes of simulated driving. We report in table \ref{tab:acc_costs} on the average costs (as defined by the matrices $\boldsymbol{Q}$ and $\boldsymbol{R}$) within the control horizon. To give a sense of the conservativeness of the car, we indicate the average velocity. 

In table \ref{tab:acc_costs}, we also compare the performance of two variations of the optimization versus the baseline: tree-2 and tree-5 have respectively 2 and 5 branches. With 80 pedestrians per km, up to 4 pedestrians can enter the planning horizon such that 5 branches are in principle, needed. Tree-2 is therefore an approximation that we use to evaluate the benefit of having larger trees versus the computation time. 
\begin{table}[h]
\begin{center}
\begin{tabular}{|c||c|c||c|c||c|}
\hline
 & \thead{pedestr.\\ per km} & \thead{percentage\\of pedestr.\\crossing} & \thead{avg.\\cost} & \thead{avg.\\speed \\ (m/s)} & \thead{planning\\time\\(ms)} \\
\hline
tree-2 & 20 & 5\% & 28.8 & 10.8 & 11.1 \\
single hyp. & 20 & 5\% & 60.3 & 8.15 & 4.92 \\
\hline
tree-2 & 20 & 25\% & 69.3 & 7.68 & 17.8 \\
single hyp. & 20 & 25\% & 83.0 & 6.18 & 7.03\\
\hline
tree-5 & 80 & 1\% & 44.5 & 8.84 & 20.4 \\
tree-2 & 80 & 1\% & 48.8 & 8.50 & 9.05 \\
single hyp. & 80 & 1\% & 86.2 & 5.45 & 3.05 \\
\hline
tree-5 & 80 & 5\% & 63.9 & 7.31 & 20.1 \\
tree-2 & 80 & 5\% & 68.3 & 6.96 & 9.61 \\
single hyp. & 80 & 5\% & 95.8 & 4.66 & 3.06 \\
\hline
tree-5 & 80 & 25\% & 109.8 & 4.18 & 19.9 \\
tree-2 & 80 & 25\% & 112.7 & 3.87 & 12.1 \\
single hyp. & 80 & 25\% & 119.5 & 3.24 & 5.52 \\
\hline
\end{tabular}
\end{center}
\caption{Performance comparison: The benefit of control-trees is greater when the likelihood of the worst case is low.}
\label{tab:acc_costs}
\end{table}

\subsubsection{Results interpretation}
Control-trees always lead to better control costs than the single hypothesis MPC (up to twice lower costs in the case of 20 pedestrians per km, and 5\% of crossing probability). In particular, the car drives less conservatively, maintains on average a higher velocity but still ensures that it is possible to come to a stop safely might a pedestrian cross. The benefit is more significant when the crossing probability is low. This can be understood easily: by discarding the probabilities, the baseline is comparatively more conservative when the probability that the worst case happens is low. 

Control-trees with 5 branches perform better than those with 2 branches only. This is consistent since the problem structure is more completely modeled with 5 branches. However, the benefit between tree-2 and tree-5 is rather small suggesting that, in this example, a simple tree is a good trade-off between performance benefit and computation time.

Control-trees take longer to optimize, but computations stay fast and compatible with a real-time application. 

\subsubsection{Scalability}
\begin{figure}[!htb]
 \center{\includegraphics[width=0.45\textwidth]{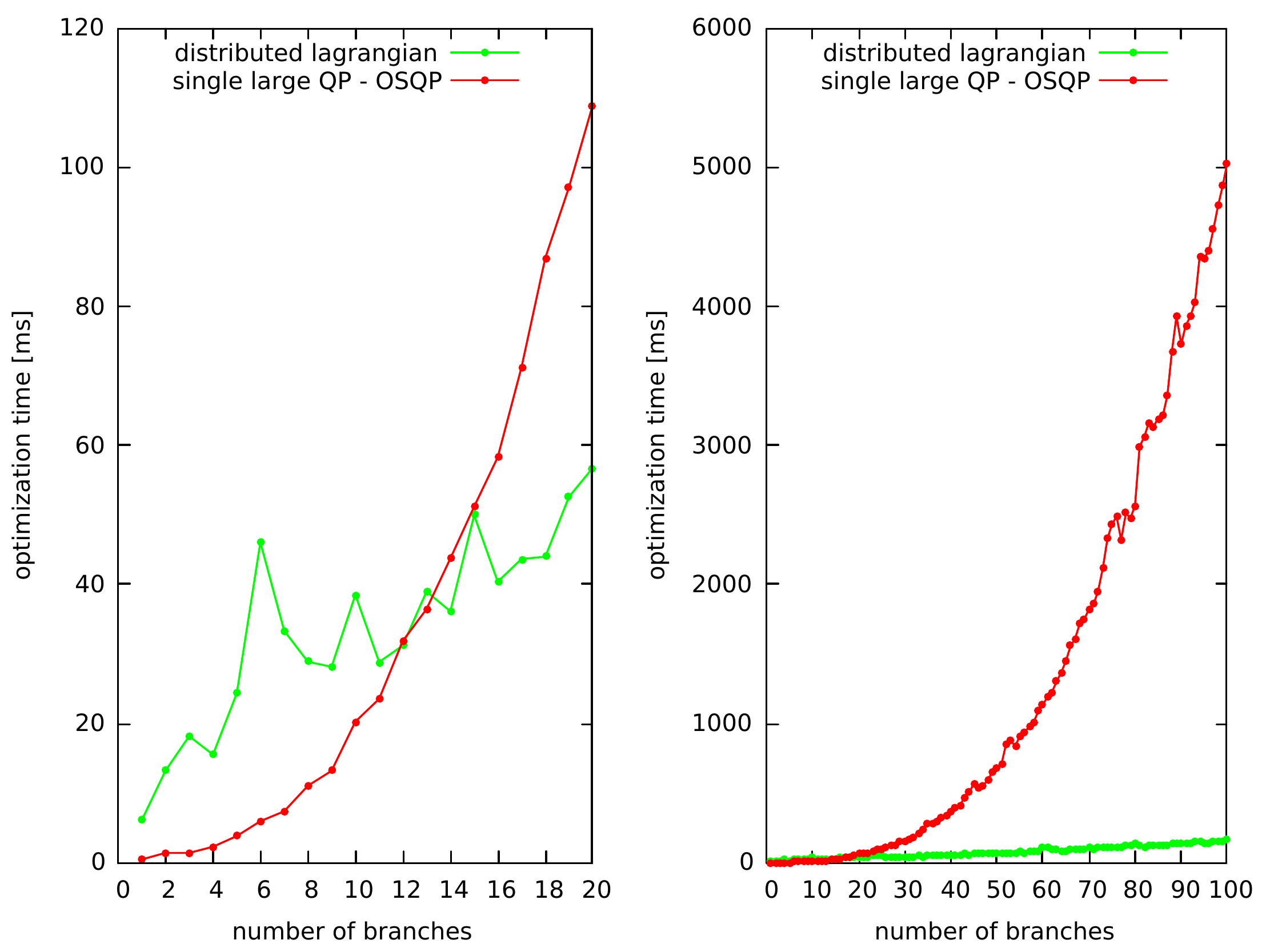}}
  \caption{Scalability: In contrast to an undecomposed optimization (red), the proposed algorithm (green) scales linearly.}
\label{fig:planning_time}
\end{figure}

We tested the optimization with up to 100 branches (see Figure \ref{fig:planning_time}). The \textit{Distributed Augmented Lagrangian} algorithm is compared to an approach where the optimization is not decomposed but cast into one large QP, and optimized with an off-the-shelf solver (OSQP see \cite{osqp}). Tests were conducted on a Intel\textsuperscript{\textregistered} Core\textsuperscript{\texttrademark} i5-8300H.

For small number of branches, our solver is slower. This reflects the fact that it is not yet, as optimized as standard solvers. However, for higher number of branches, the algorithm is drastically more efficient than the undecomposed optimization. In practice, it scales linearly with the number of hypotheses. Indeed, the number of subproblems increases with the number of hypotheses (since they are equal), but the size of each subproblem doesn't change.

\subsection{Slalom among uncertain obstacles}  \label{section:slalom}
In this example, the ego-vehicle drives along a straight reference trajectory. Obstacles appear randomly. Obstacle detection is imperfect: there are \textit{false-positives} i.e. obstacles are detected although they don't exist. This captures a common problem when working with sensors like radars that can suffer from a high rate of false detections. The simulated detection module also outputs an existence probability for each detected object. The closer the car gets to the obstacle, the more reliable are the observations. Once the distance to obstacle becomes lower than a threshold (randomized in the simulation), the object becomes fully observed: it gets a probability of 1.0 or disappears.

Optimization is performed in configuration space using the \textit{K-Order-Motion-Optimization} formulation (KOMO, see \cite{komo-2}\cite{komo-1}). The horizon is set to \SI{5}{\second} with 4 steps per second.
\begin{figure}[ht]
        \centering
  \centering
  \includegraphics[width=0.95\linewidth]{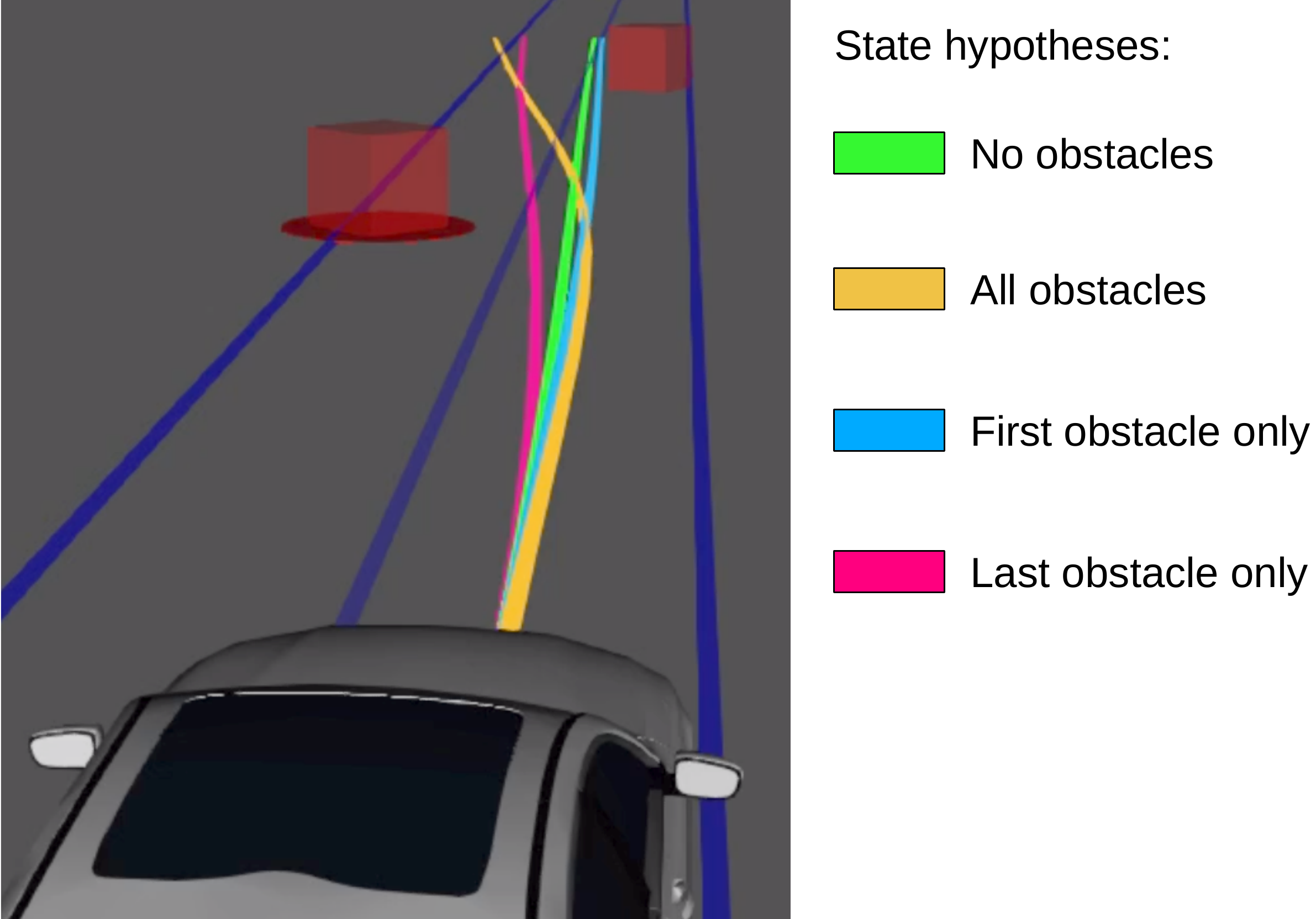}  
\caption{Avoidance of 2 uncertain obstacles: Each trajectory corresponds to a different combination of object presences.}
\label{fig:avoidance_tree_vs_linear}
\end{figure}



\subsubsection{State space and kinematic} \label{section:non-holonomic}
Optimization is performed in $SE(2)$. No slippage is assumed, this is enforced by the following non-holonomic constraint:
\begin{subequations}
\begin{align}
 \dot{x}(t) cos(\theta) - \dot{y}(t) \sin(\theta) = 0, \forall{t}, \label{eq:non-holonomic}
\end{align}
\end{subequations}
where $\begin{pmatrix} x, y, \theta \end{pmatrix} $ is the vehicle pose.

\subsubsection{Partially observable discrete state}
The discrete states represent the possible combinations of obstacle existences. With 2 uncertain obstacles in the planning horizon, there are 4 possible states (see Figure \ref{fig:avoidance_tree_vs_linear}).

\subsubsection{Costs and constraints}
The following trajectory cost terms are minimized:
\begin{itemize}
\item \textbf{Acceleration}: The square accelerations $\ddot{x}$ and $\ddot{y}$ as well as the square angular acceleration $\ddot{\theta}$.
\item \textbf{Distance to centerline}: The square distance to a reference line.
\item \textbf{Speed}: The square difference to a desired velocity.
\end{itemize}

\noindent In addition, the following constraints apply:
\begin{itemize}
\item \textbf{Kinematic}: The non-holonomic constraint (see (\ref{eq:non-holonomic})).
\item \textbf{Collision Avoidance}: The distance to obstacle must stay greater than a safety distance.
\end{itemize}

Unlike the previous example, the non-holonomic and collision avoidance constraints make the problem \textit{non-convex}.

Function gradients are computed analytically while we use the gauss-newton approximation of the \nth{2} order derivatives.


\subsubsection{Evaluation on random scenarios}
Table \ref{tab:table_obstacle_avoidance} gathers results obtained when simulating 30 minutes of driving, with on average, one potential obstacle every 17 meters resulting overall in 900 uncertain obstacles being encountered. As in the previous example, the baseline used here, is a single-hypothesis MPC which doesn't use the belief state information: all uncertain obstacles are avoided the same way, regardless of their existence probabilities.
\begin{table}[h]
\begin{center}
\begin{tabular}{|c||c|c||c|c||c|}
\hline
  & \thead{percentage of \\ false positive} & \thead{avg.\\cost} & \thead{planning\\time (ms)} \\
\hline
tree-4  & 90\% & 0.42 & 59.3 \\
single hyp. & 90\% & 0.91 & 27.7 \\
\hline
tree-4  & 75\% & 0.97 & 74.2 \\
single hyp. & 75\% & 1.8 & 32.8 \\
\hline
\end{tabular}
\end{center}
\caption{Performance comparison: Control-trees lead to lower costs.}
\label{tab:table_obstacle_avoidance}
\end{table}

Compared to the baseline, the control-tree consistently provides lower trajectory costs within the control horizon (approximately twice lower). In particular, the vehicle experiences less lateral acceleration while still avoiding successfully all real obstacles.

It is also interesting to compare those costs to the ideal fully observable case i.e. where the agent has perfect information and knows which obstacles are real, and which are false detections. This ideal case gives a lower bound of the trajectory costs which amount to $0.3$ and $0.78$ for the scenarios with $90\%$ and $75\%$ of false detections respectively. 
With this point of comparison, the benefits compared to the baseline is even more significant.

Computation time increases but remains largely compatible with the planning frequency of 10Hz.

\section{Conclusion and future work}
We proposed a new MPC approach for problems where discrete and critical aspects of the world are partially observed. Optimization is performed using probabilistic information, which results in policies that outperform policies optimized under a single hypothesis MPC scheme. 

An optimization algorithm, \textit{Distributed Augmented Lagrangian} is proposed which leverages the decomposable structure of the optimization problem to improve scalability.

In future work, our ambition is to scale the approach to scenarios requiring an order of magnitude more states. As this may not be possible to do so in real time, we would like to explore ways to use this multi-hypotheses MPC framework to compute offline, expert demonstrations that can be used to learn control policies.


\bibliography{references}
\bibliographystyle{ieeetr}

\end{document}